# Generating Thermal Image Data Samples using 3D Facial Modelling Techniques and Deep Learning Methodologies

Muhammad Ali Farooq
*College of Engineering and Informatics*
*National University of Ireland Galway (NUIG)*
Galway, Ireland
m.farooq3@nuigalway.ie

Peter Corcoran
*College of Engineering and Informatics*
*National University of Ireland Galway (NUIG)*
Galway, Ireland
peter.corcoran@nuigalway.ie

*Abstract*— **Methods for generating synthetic data have become of increasing importance to build large datasets required for Convolution Neural Networks (CNN) based deep learning techniques for a wide range of computer vision applications. In this work, we extend existing methodologies to show how 2D thermal facial data can be mapped to provide 3D facial models. For the proposed research work we have used tufts datasets for generating 3D varying face poses by using a single frontal face pose. The system works by refining the existing image quality by performing fusion based image preprocessing operations. The refined outputs have better contrast adjustments, decreased noise level and higher exposedness of the dark regions. It makes the facial landmarks and temperature patterns on the human face more discernible and visible when compared to original raw data. Different image quality metrics are used to compare the refined version of images with original images. In the next phase of the proposed study, the refined version of images is used to create 3D facial geometry structures by using Convolution Neural Networks (CNN). The generated outputs are then imported in blender software to finally extract the 3D thermal facial outputs of both males and females. The same technique is also used on our thermal face data acquired using prototype thermal camera (developed under Heliaus EU project) in an indoor lab environment which is then used for generating synthetic 3D face data along with varying yaw face angles and lastly facial depth map is generated.**

I. INTRODUCTION

With the recent advancements in technology and the growing requirements for larger datasets, it is very important to extract maximum information from the acquired data. Visible or RGB data is most commonly used for a wide range of computer vision applications however it is not able to generate temperature patterns of the specific body which is an important factor in critical applications. Thermal cameras can capture the temperature patterns of the human body by sensing the emission of infrared radiation. Thermal data of the human body is considered to be very important for many applications such as human disease diagnosis in early stages by extracting human facial and body temperature patterns and medical image analysis techniques and creating 3D synthetic thermal face data for visualization and animations. In the proposed study we have proposed 3D synthetic thermal face data generation by using advanced deep learning methods inspired by Feng, Yao, et al [1]. Such types of data can be used in different types of human biometric applications such as thermal facial recognition systems, gender classification system, emotion recognition systems.

As compared to 2D facial images 3D facial structures can help in dealing with the problem of varying human poses and occlusions. Deep learning algorithms have played a vital role in solving many computer vision applications including 3D data creation by taking advantage of convoluting neural networks (CNN). CNN is well known for its self-feature extraction from the raw pixel of images rather than relying on handcrafted features which are subsequently required for conventional machine learning algorithms.

The rest of the paper is structured as follows, section II provides the background and related research for creating synthetic data, section III describes the proposed methodology regarding image refinement and deep learning for generating the 3D facial structures, section IV provides experimental results and lastly, section V describes the conclusion and future work.

II. Background

Synthetic data can be considered as a repository of data that is not collected from real-world experiments, but it is generated programmatically by using different algorithms and methodologies from the domain of machine learning and pattern recognition. The most common approach for generating synthetic data is to pick the work of 3D artists they have done by creating real-time virtual environments for video gaming. Richter et al. [2] have captured datasets from the Grand Theft Auto V video game. The authors mainly emphasize on using semantic segmentation methods. Authors have captured the communication between the game and the graphics hardware. Through this approach, they have cut the labeling costs (in annotation time). Qiu et al. [4] have developed UnrealCV which is an open-source plugin for the popular game engine Unreal Engine 4. It works by providing commands that allow us to get and set camera location, field of view and get the set of objects in a scene together with their positions. Choi et al. [6] created a dataset of 3D models of household objects for their tracking filter. Moreover, Hodan et al. [7] provide a real dataset of textureless objects supplemented with 3D models of these objects. Deep learning has been extensively used for generating 3D data especially biometrics data for various real-world applications. Dou et al. [8] and Tuan Tran, Anh, et al [9] have proposed an end to end CNN architectures to directly estimate the 3D morphable models (3DMM) shape parameters. In this study, we have proposed a novel method to generate a comprehensive 3D thermal facial structure by using state of art CNN architecture.

III. PROPOSED METHODOLOGY

This section of the paper will mainly focus on the proposed algorithm for generating 3D synthetic face data from a single 2d thermal image. We have utilized the tufts thermal face dataset [10] since this dataset has published recently with data samples from 6 different image modalities which include visible, near-infrared, thermal, computerized sketch, a recorded video, and 3D images. The dataset consists of images of both males and females genders. It was acquired in an indoor environment using FLIR Vue Pro Camera with constant lighting. Fig. 1 displays sample thermal images of both male and female subjects from the tufts dataset.

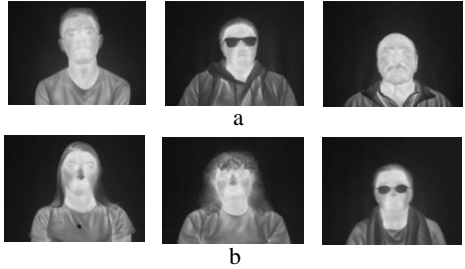

Fig. 1. Thermal face images from tufts dataset a) male samples, b) female samples

In the first phase, the system works by taking a single frontal pose and producing the refined version of the image to make facial features such as eyes, lips, nose and temperature patterns on the face more visible and vibrant. This approach works by applying various image preprocessing operations built on multi-scale fusion principles inspired by Ancuti, Cosmin, et al [13].

A. 2D Image Processing

The algorithm consists of six main steps. In the first step, algorithms work by applying a simple white color balance which is color corrections operation to remove the unlikely color casts in order to render specific colors in an image. The resulting outputs are color corrected version with reduced noise levels as compared to original raw data. In the second stage,

Contrast limited Adapt Histogram Equalization (CLAHE) is applied to enhance the visibility of confined details by improving the contrast of local regions in the image. It is done to achieve optimal contrast levels in the input image. In the next stage, different types of weighs are applied to increase the exposure levels in the dark regions. This is achieved by applying four different types of weights which include laplacian contrast weights, local contrast weights, saliency contrast weights, and exposedness weights [13]. Laplacian weights are generally used to enhance the global contrast of the image. Local contrast weights are applied to strengthen the local contrast appearance since it advantages the transitions mainly in the highlighted and shadowed parts of images. It is computed as the standard deviation between pixel luminance level and the local average of its surrounding region as shown in equation 1 [12].

$$W_{lc}(x,y) = II(I^k - I^k_{Whc})II \quad (1)$$

Where $W_{lc}(x,y)$ represent the symbol for local contrast weights, $I^k$ represents the luminance channel of the input and the $I^k_{Whc}$ represents the low-passed version of it.

Saliency weights are applied to emphasize the discriminating objects that lose their prominence especially in the dark regions and exposedness weights are finally applied to evaluate how well the pixel is exposed. The weights are measured by using the saliency algorithm of Achanta et al. [1]. Fig. 2 represents the complete multi-scale fusion image refinement process on the thermal frontal face image from the tufts dataset.

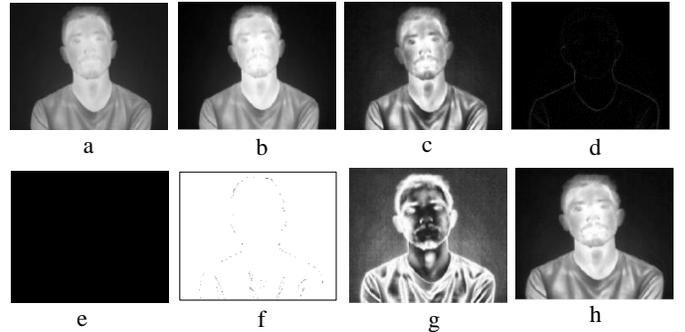

Fig. 2. Complete multi-scale image fusion algorithm pipeline to produce refined version of images a) input image, b) white color balance applied, c) histogram equalization applied, d) laplacian contrast weight applied, e) local contrast weights applied, f) saliency weights applied, g) exposedness weights applied, h) final output image

B. 2D to 3D Image Reconstruction

Once the images are refined in the second stage, we have used end to end convolution neural network also referred to as Position Map Regression Network (PRN) [1] to reconstruct the 3D images from a single frontal face pose thermal image. The authors in [1] had proposed CNN network which was trained to generate 3D facial structures using one single RGB image.

The network works by transferring the input image into a position map. In the next stage, the authors have used the encoder-decoder structure for learning the transfer structure.

This project (https://www.heliaus.eu/) has received funding from the ECSEL Joint Undertaking (JU) under grant agreement No 826131. The JU receives support from the European Union's Horizon 2020 research and innovation program and France, Germany, Ireland, Italy.

The encoder structure is consisting of one convolution layer which is followed by a series of ten residual blocks for performing downsampling operation. The decoder structure is consisting of seventeen transposed convolutions blocks in order to generate the predicted output position map. The proposed CNN networks use Rectified linear Unit (ReLU) activation functions and a kernel size of four is used for each of the convolution layers and transposed convolution layers. A customized loss function was built to learn the parameters to a better extent by measuring the difference between the ground truth position map and the network output. The loss function utilizes the weight mask which is the grey image recording the weight of each map on the position map. The weigh mask is of the same size and pixel to pixel correspondence when compared with the position map. The loss function is defined in equation 2 [1].

$$Loss = \sum \| Pos(u,v) - Pos\sim(u,v) \| \cdot W(u,v) \quad (2)$$

Where $Pos(u,v)$ represent predicted position map, $Pos\sim(u,v)$ represents ground truth position map and $W(u,v)$ represents the weight mask.

In this study, we have used the same network for generating synthetic 3D facial geometry structures using one single image. However, instead of using the visible image we have utilized thermal facial images to validate the effectiveness of the network. The network initially produces .obj file which is then imported in blender software [14] to generate the 3D facial geometry as the output. The complete workflow diagram is shown in Fig. 3.

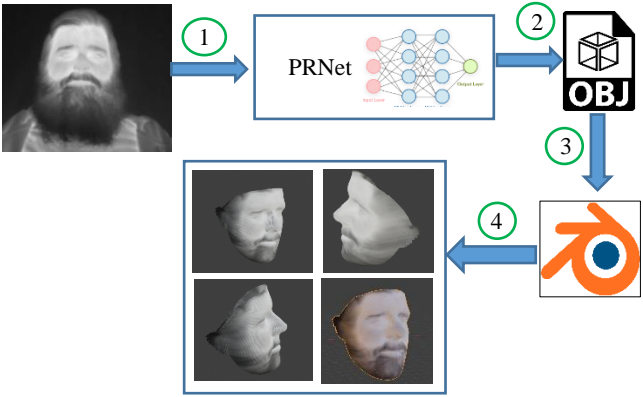

Fig. 3. Complete workflow diagram for generating the synthetic 3D facial structure from single 2D thermal image 1: input image fed to PRNet for generating 3D facial data, 2: output obj file, 3: obj file imported to blender software, 4: final outputs extracted in the form of 3D thermal facial images covering different poses

## IV. EXPERIMENTAL RESULTS

The overall algorithm was implemented using Matlab R2018a for applying fusion based image preprocessing methods as discussed in Section III to produce the refined version of images. TensorFlow [15] deep learning framework was used for generating 3D facial structures using the pre-trained PRNet [1]. The system was deployed and tested on the Core I7 machine with 32 GB of RAM equipped with NVIDIA RTX 2080 Graphical Processing Unit (GPU) having 8GB of dedicated graphic memory.

The first phase of the experimental results shows the refined outputs obtained by applying fusion based image preprocessing operations. It makes the facial features and temperature patterns of the face more visible by reducing the overall noise and adjusting the optimal brightness and contrast levels in the image. It is shown in Fig. 4.

In the proposed study we have used different image quality metrics on both processed thermal as well normal dataset images which include, Naturalness Image Quality Evaluator (NIQE) [16] and Blind/Reference less Image Spatial Quality Evaluator (BRISQE) [17] score to compare the enhanced version of images with original (ground-truth) images. It is shown in Table TABLE I.

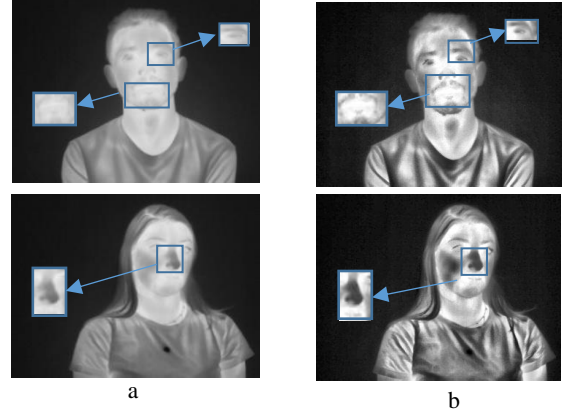

a           b

Fig. 4. Image preprocessing results a) input images of two different subject (male and female), b) refined outputs with more visible facial features

TABLE I.     IMAGE QUALITY METRICS

| Image | NIQE Score | BRISQUE Score |
|---|---|---|
| Processed Image | 3.0323 lower is better | 18.2226 lower is better |
| Orignal Image | 3.3350 | 37.2575 |
| Processed Image | 2.9099 lower is better | 29.4339 lower is better |
| Orignal Image | 3.6029 | 35.7555 |
| Processed Image | 2.7059 lower is better | 12.0921 lower is better |
| Orignal Image 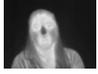 | 3.4460 | 33.5439 |

The lower NIQE AND BRISQUE scores of processed images evidently reflect that image quality is improved significantly by applying the image refinement/ prepossessing techniques. The second phase of the experimental results demonstrates the 3D facial structures generated through PRNet and extracted through blender software using a refined version of the thermal image. It is shown in Fig. 5.

The same process was used to produce varying 3D facial poses of both male and female samples from the tufts dataset. It is shown in Fig. 6.

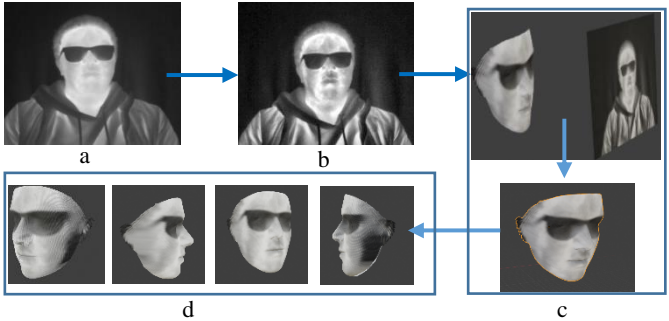

Fig. 5. 3D synthetic face structure of male sample a) input image, b) refined/preprocessed image, c) 3D face geometry generated in blender software, d) different face poses of the subject

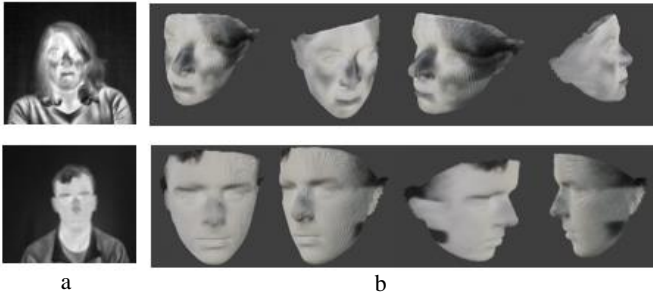

Fig. 6. 3D synthetic face structure of both male and female gender a) refined input images of female and male, b) synthetic 3D outputs with varying face poses

Further experimental results demonstrate 3D facial structures using our own dataset. The data was gathered in the indoor lab environment using a prototype thermal camera that embeds a Lynred [21] LWIR sensor developed under the Heliaus EU project [20]. Fig. 7 displays the prototype thermal camera model being used for the proposed research work whereas Table TABLE II. provides the technical specifications of the camera.

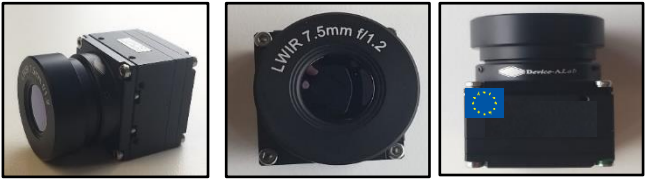

Fig. 7. Prototype thermal VGA camera model for acquiring local data

TABLE II. TECHNICAL SPECIFICATIONS OF PROTOTYPE THERMAL CAMERA

| | Specifications |
|---|---|
| Type | Long Wave Infrared |
| Resolution | 640 x 480 pixels |
| Quality | VGA |
| Focal length | 7.5 mm |

The data was collected by mounting the camera on a tripod stand with the fixed distance (nearly 60 centimeters) from the subject The data acquisition structure is shown in Fig. 8.

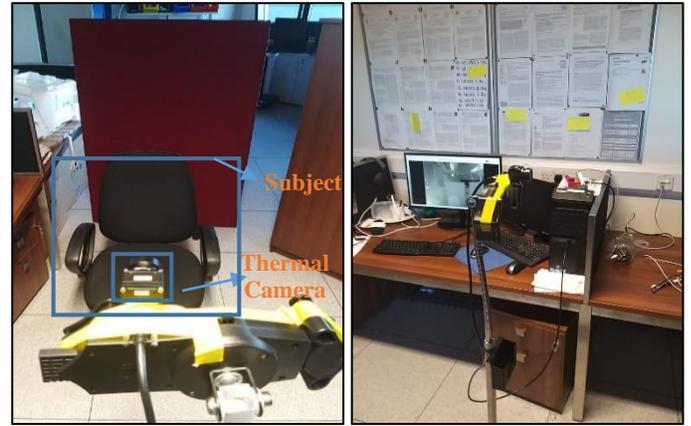

Fig. 8. Data acquisition setup in indoor university lab environment, smart R640L thermal camera mounted on tripod stand placed at a fixed distance (60 centimeters) from the subject

We have collected data in two different modalities which include RGB and thermal respectively. For preliminary testing, only male subjects took part in this study. The data is gathered by recording the video and then extracting the image sequence from the video. Fig. 9 displays frontal face poses of two different subject referred to as subject A and subject B who have taken part in this study along with their processed thermal outputs.

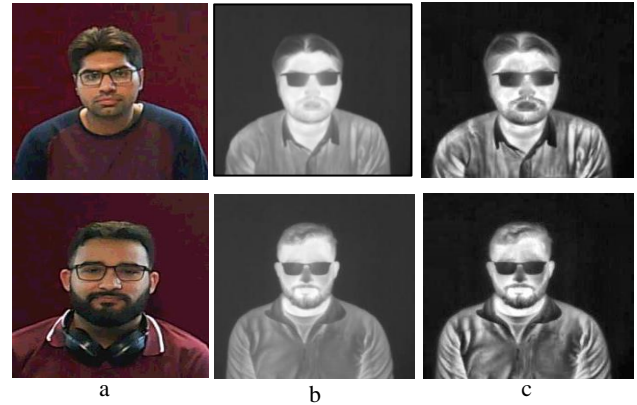

Fig. 9. Face data samples of two different subjects acquired in an indoor lab environment a) visible image, b) thermal image, c) processed thermal image

After collecting the data, the same technique was used to generate 3D facial structures. Along with 3D facial structures, we have also generated a facial depth map. It is exhibited in Fig. 10.

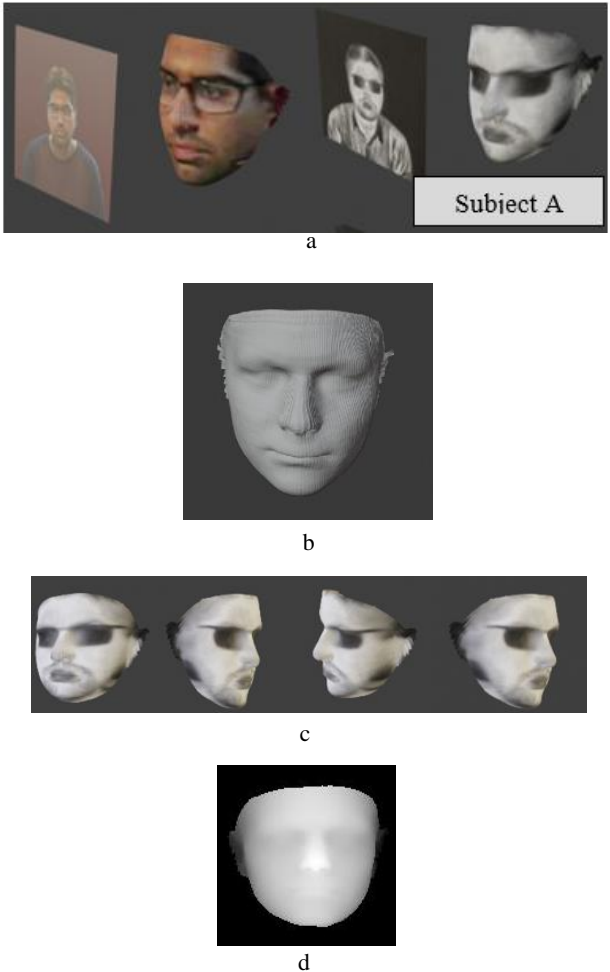

Fig. 10. Synthetic 3D facial structure results with depth map on our own data a) RGB and thermal Obj file imported in Blender software, b) 3D facial mesh , c) different face yaw angles, d) face depth map

## V. CONCLUSION AND FUTURE WORK

In the proposed study we have incorporated advance deep learning model PRNet [1] for generating synthetic 3D thermal facial structures from the single thermal frontal image. Such type of data can be found useful in a variety of real-time computer vision and machine learning applications such as medical image analysis, extracting facial and body temperatures for in-cabin driver monitoring systems, visualization, and animation creation. As compared to conventional data generation techniques such as data augmentation and data transformation that can produce 2D outputs, 3D synthetic data can be found more robust and realistic especially when training deep neural networks for critical applications. Along with thermal and visible data, we can use this methodology to create synthetic data from other image modalities such as infrared, near-infrared and grayscale data. Moreover, as future work, we can train CNN networks to generate 3D facial textures that can be aligned with 3D mesh for generating animations, different facial expressions and facial emotions with varying lighting conditions, and occlusions.

ACKNOWLEDGMENT

This research is supported and funded by the Heliaus European Union Project. The project focused on enabling safe autonomous driving systems. This project has received funding from the ECSEL Joint Undertaking (JU) under grant agreement No 826131. The JU receives support from the European Union's Horizon 2020 research and innovation program and France, Germany, Ireland, Italy. The authors would like to acknowledge Shubhajit Basak for providing his support to use blender software, the Xperi Ireland team and Quentin Noir from Lynred France for giving their feedback. Moreover, authors would like to acknowledge tufts university the contributors of the tufts dataset for providing the image resources to carry out this research work.

REFERENCES

[1] Feng, Yao, et al. "Joint 3d face reconstruction and dense alignment with position map regression network." Proceedings of the European Conference on Computer Vision (ECCV). 2018.

[2] Stephan R. Richter, Zeeshan Hayder, and Vladlen Koltun. Playing for benchmarks. CoRR, abs/1709.07322, 2017.

[3] Stephan R. Richter, Vibhav Vineet, Stefan Roth, and Vladlen Koltun. Playing for data: Ground truth from computer games. CoRR, abs/1608.02192, 2016.

[4] Weichao Qiu and Alan L. Yuille. Unrealcv: Connecting computer vision to unreal engine. CoRR, abs/1609.01326, 2016.

[5] Weichao Qiu, Fangwei Zhong, Yi Zhang, Zihao Xiao Siyuan Qiao, Tae Soo Kim, Yizhou Wang, and Alan Yuille. Unrealcv: Virtual worlds for computer vision. ACM Multimedia Open Source Software Competition, 2017.

[6] C. Choi and H. I. Christensen. Rgb-d object tracking: A particle filter approach on gpu. In 2013 IEEE/RSJ International Conference on Intelligent Robots and Systems, pages 1084–1091, Nov 2013.

[7] Tomas Hodan, Pavel Haluza, Step´an Obdrz´alek, Jiri Matas, Manolis I. A. Lourakis, and Xenophon Zabulis. T-LESS: an RGB-D dataset for 6d pose estimation of texture-less objects. CoRR, abs/1701.05498, 2017.

[8] Dou, Pengfei, Shishir K. Shah, and Ioannis A. Kakadiaris. "End-to-end 3D face reconstruction with deep neural networks." Proceedings of the IEEE Conference on Computer Vision and Pattern Recognition. 2017.

[9] Tuan Tran, Anh, et al. "Regressing robust and discriminative 3D morphable models with a very deep neural network." Proceedings of the IEEE Conference on Computer Vision and Pattern Recognition. 2017.

[10] Tufts Thermal Dataset, Weblink: http://tdface.ece.tufts.edu/, Last accessed on 29 October 2019.

[11] Panetta, Karen, Qianwen Wan, Sos Agaian, Srijith Rajeev, Shreyas Kamath, Rahul Rajendran, Shishir Rao et al. "A comprehensive database for benchmarking imaging systems." IEEE Transactions on Pattern Analysis and Machine Intelligence (2018).

[12] Paper: Shreyas Kamath K. M., Rahul Rajendran, Qianwen Wan, Karen Panetta, and Sos S. Agaian "TERNet: A deep learning approach for thermal face emotion recognition", Proc. SPIE 10993, Mobile Multimedia/Image Processing, Security, and Applications 2019, 1099309 (13 May 2019).

[13] Ancuti, Cosmin, et al. "Enhancing underwater images and videos by fusion." 2012 IEEE Conference on Computer Vision and Pattern Recognition. IEEE, 2012.

[14] Blender software Website: https://www.blender.org/, (Last accessed on 31st December 2019).

[15] TensorFlow Deep Learning Platform, Website: https://www.tensorflow.org, (Last accessed on 29th December 2019).

[16] Mittal, Anish, Anush Krishna Moorthy, and Alan Conrad Bovik. "No-reference image quality assessment in the spatial domain." IEEE Transactions on image processing 21.12 (2012): 4695-4708.

[17] Mittal, Anish, Rajiv Soundararajan, and Alan C. Bovik. "Making a "completely blind" image quality analyzer." IEEE Signal Processing Letters 20.3 (2012): 209-212.

[18] Huynh-Thu, Quan, and Mohammed Ghanbari. "Scope of validity of PSNR in image/video quality assessment." Electronics letters 44.13 (2008): 800-801.

[19] Cheng, Ming-Ming, et al. "Global contrast based salient region detection." IEEE Transactions on Pattern Analysis and Machine Intelligence 37.3 (2014): 569-582.

[20] Heliaus European Union Project Website: https://www.heliaus.eu/, (Last accessed on 20th January 2020).

[21] Lynred France Website: https://www.lynred.com/, (Last accessed on 27th January 2020).